\newif\iftodo
\todotrue 

\newif\ifjournal\journaltrue

\ifjournal
    \documentclass{IEEEtran}
\else

    \documentclass[12pt,journal, onecolumn]{IEEEtran}
	\usepackage[english]{babel} 

\fi

\usepackage{graphicx}
\usepackage{amsfonts}
\usepackage{subfigure}
\usepackage{float}
\usepackage{bm}
\usepackage{algorithm}
\usepackage{algorithmic}
\usepackage{amsmath}
\usepackage{amsthm}
\usepackage{caption}  
\usepackage{booktabs} 
\usepackage{multirow}
\usepackage{setspace}
\usepackage{tabularx}

\usepackage{xr}

\usepackage[table]{xcolor}
\newcolumntype{s}{>{\columncolor[HTML]{FE6F5E}} c}
\newcolumntype{t}{>{\columncolor[HTML]{5D8AA8}} c}

\usepackage[normalem]{ulem}                       
\newcommand\redout{\bgroup\markoverwith{\textcolor{red}{\rule[0.5ex]{2pt}{0.8pt}}}\ULon}

\usepackage[square,numbers,sort&compress]{natbib} 
\usepackage[capitalise]{cleveref}                 
\crefname{equation}{}{}                           
\Crefname{equation}{}{}                           

\iftodo
  \usepackage[colorinlistoftodos]{todonotes}
\else
  \usepackage[colorinlistoftodos,disable]{todonotes}
\fi
\makeatletter
\if@todonotes@disabled
\else
    \usepackage[left=1.8cm, right=1.8cm, top = 1.8cm, bottom=1.5cm, letterpaper]{geometry}
    \newdimen\extramargin
\fi
\makeatother

\newcommand{\Tscr}{\mathcal T}

\newcommand{\bbR}{\mathbb R}

\newcommand{\prox}{\textrm{prox}}

    \DeclareMathOperator*{\argmin}{\arg\!\min}

    \renewcommand{\v}[1]{{\bf{#1}}}

\newcommand{\etal}{\textit{et al.}\,}

\usepackage[]{breqn}

\title{Deep Unfolded Robust PCA with Application to Clutter Suppression in Ultrasound}

\author{Oren Solomon,~\IEEEmembership{Student Member,~IEEE,} Regev Cohen,~\IEEEmembership{Student Member,~IEEE,}
\thanks{O. Solomon and R. Cohen contributed equally to this work.}
Yi Zhang, Yi Yang, He Qiong, Jianwen Luo, Ruud J.G. van Sloun,~\IEEEmembership{Member,~IEEE},   
and Yonina C. Eldar,~\IEEEmembership{Fellow,~IEEE} 
    \thanks{This project has received funding from the European Union's Horizon 2020 research and innovation program under grant agreement No. 646804-ERC-COG-BNYQ.}
\thanks{O. Solomon (e-mail:  orensol@campus.technion.ac.il), R. Cohen (e-mail:  regev,cohen@campus.technion.ac.il) and Y. C. Eldar (e-mail: yonina@ee.technion.ac.il) are with the Department of Electrical Engineering, Technion-Israel Institute of Technology, Haifa 32000.
Y. Zhang (e-mail: yizhang.ch.2015@gmail.com) is with the department of electrical engineering, Tsinghua University, Beijing 100084, China.
Y. Yang, Q. He and J. Luo are with the Department of Biomedical Engineering, Tsinghua University, Beijing 100084, China.
R. J. G. van Sloun (e-mail:  R.J.G.v.Sloun@tue.nl) is with the Department of Electrical Engineering, Eindhoven University of Technology, Eindhoven, The Netherlands.}
}

\begin{document}
\maketitle

\begin{abstract}
Contrast enhanced ultrasound is a radiation-free imaging modality which uses encapsulated gas microbubbles for improved visualization of the vascular bed deep within the tissue. It has recently been used to enable imaging with unprecedented subwavelength spatial resolution by relying on super-resolution techniques. A typical preprocessing step in super-resolution ultrasound is to separate the microbubble signal from the cluttering tissue signal. This step has a crucial impact on the final image quality. 
Here, we propose a new approach to clutter removal based on robust principle component analysis (PCA) and deep learning. We begin by modeling the acquired contrast enhanced ultrasound signal as a combination of a low rank and sparse components. This model is used in robust PCA and was previously suggested in the context of ultrasound Doppler processing and dynamic magnetic resonance
imaging. We then illustrate that an iterative algorithm based on this model
exhibits improved separation of microbubble signal from the tissue
signal over commonly practiced methods. Next, we apply the concept of deep unfolding to suggest a deep network architecture tailored to our clutter filtering problem which exhibits improved convergence speed and accuracy
with respect to its iterative counterpart. We compare the performance of
the suggested deep network on both simulations and in-vivo rat brain
scans, with a commonly practiced deep-network architecture and
the fast iterative shrinkage algorithm, and show that
our architecture exhibits better image quality and contrast.
\end{abstract}


\section{Introduction}
\IEEEPARstart{M}{edical} ultrasound (US) is a radiation-free imaging modality used extensively for  diagnosis in a wide range of clinical segments such as radiology, cardiology, vascular, obstetrics and emergency medicine. Ultrasound-based imaging modalities include brightness, motion, Doppler, harmonic modes, elastography and more \cite{fenster2015ultrasound}.

One important imaging modality is contrast-enhanced ultrasound (CEUS) \cite{furlow2009contrast} which allows the detection and visualization of blood vessels whose physical parameters such as relative blood volume (rBV), velocity, shape and density are associated with different clinical conditions \cite{opacic2018motion}. CEUS uses encapsulated gas microbubbles as ultrasound contrast agents (UCAs) which are administrated intravenously and are similar in size to red blood cells and thus can flow throughout the vascular system \cite{de1991principles}. 
Among its many applications, CEUS is used for imaging of perfusion at the capillary level \cite{lassau2007dynamic,hudson2015dynamic}, for estimating blood velocity in small vessels arteriole by applying Doppler processing \cite{tremblay2014combined,tremblay2016visualizing} and for sub-wavelength vascular imaging \cite{bar2017sparsity,van2017sparsity,van2018super,solomon2018exploiting,Errico2015,christensen2015vivo}.

A major challenge in ultrasonic vascular imaging such as CEUS is to suppress clutter signals stemming from stationary and slowly moving tissue as they introduce significant artifacts in blood flow imaging \cite{bjaerum2002clutter}. Over the past few decades several approaches have been suggested for clutter removal. The simplest method to remove tissue signal is to filter the 
ultrasonic signal along the temporal dimension using high-pass finite impulse response (FIR) or infinite impulse response (IIR) filters \cite{thomas1994improved}. 
However, FIR filters need to have high order while IIR filters exhibit a long settling time which leads to a low number of temporal samples in each spatial location \cite{yoo2003adaptive} 
when using focused transmission. 
The above methods rely on the assumption that tissue motion, if exists, is slow while blood flow is fast. This high-pass filtering approach is prone to failure in the presence of fast tissue motion, as in cardiology, or when imaging microvasculature in which blood velocities are low. 

An alternative method for tissue suppression is second harmonic imaging \cite{frinking2000ultrasound}, which separates the blood and tissue signals by exploiting the non-linear response of the UCAs to low acoustic pressures, compared with the mostly linear tissue response. This technique, however, limits the frame-rate of the US scanner, and does not remove the tissue signal completely, as tissue can also exhibit a nonlinear response.

The above techniques are based only on temporal information and neglect the high spatial coherence of the tissue, compared to the blood. To take advantage of these spatial characteristics of tissue, a method for clutter removal was presented in \cite{ledoux1997reduction}, based on the singular value decomposition (SVD) of the correlation matrix of successive temporal samples. 
SVD filtering operates by stacking the (typically beamformed) acquired frames as vectors in a matrix whose column index indicates frame number. Then, an SVD of the matrix is performed and the largest singular values, which correspond to the highly correlated tissue, are zeroed out. Finally, a new matrix is composed based on the remaining singular values and reshaped to produce the blood/UCA movie. 

Several SVD-based techniques have been proposed \cite{yu2010eigen,bjaerum2002clutter,mauldin2010complex,mauldin2011singular,gallippi2003bss}
, such as down-mixing \cite{bjaerum2002clutter} for tissue motion estimation, adaptive clutter rejection for color flow proposed by Lovstakken \etal \cite{lovstakken2006real} and the principal component analysis (PCA) for blood velocity estimation presented in \cite{kruse2002new}. However, these methods are based on focused transmission schemes which limit the frame rate and the field of view. This in turn leads to a small number of temporal and spatial samples, reducing the effectiveness of SVD-based filtering. To overcome this limitation, SVD-based clutter removal was extended to ultrafast plane-wave imaging \cite{demene2015spatiotemporal,Errico2015,song2017ultrasound,chee2018receiver},  demonstrating substantially improved clutter rejection and microvascular extraction. This strategy gained a lot of interest in recent years and nowadays it is used in numerous ultrafast US imaging applications
such as functional ultrasound \cite{urban2015real,errico2016transcranial}, super-resolution ultrasound localization microscopy \cite{Errico2015,christensen2015vivo} and high-sensitivity microvessel perfusion imaging \cite{demene2015spatiotemporal,song2017ultrasound}. 

A major limitation of SVD-based filtering is the requirement to determine a threshold which discriminates between tissue related and blood related singular values. The appropriate setting of this threshold is typically unclear, especially when the eigenvalue spectra of the tissue and contrast signals overlap. This threshold uncertainty motivates the use of a different model for the acquired data, one that can differentiate between tissue and contrast signals based on the spatio-temporal information, as well as additional information unique to the contrast signal - its sparse distribution in the imaging plane.   


Here, we propose two main contributions. The first, is the adaptation of a new model for the tissue/contrast separation problem. 
We show that similar to other applications such as MRI \cite{otazo2015low} and recent US Doppler applications \cite{bayat2018concurrent}, we can decompose the acquired, beamformed US movie as a sum of a low-rank matrix (tissue) and a sparse outlier signal (UCAs). This decomposition is also known as robust principle component analysis (RPCA) \cite{candes2011robust}. We then propose to solve a convex minimization problem to retrieve the UCA signal, which leads to an iterative principal component pursuit (PCP) \cite{candes2011robust}. Second, we utilize recent ideas from the field of deep learning \cite{lecun2015deep} to dramatically improve the convergence rate and image reconstruction quality of the iterative algorithm. We do so by unfolding \cite{gregor2010learning} the algorithm into a fixed-length deep network which we term Convolutional rObust pRincipal cOmpoNent Analysis (CORONA).
This approach harnesses the power of both deep learning and model-based frameworks, and leads to improved performance in various fields \cite{sprechmann2015learning,sreter2018learned,giryes2018tradeoffs,giryes2018learned,neev2018detect}. 

CORONA is trained on sets of separated tissue/UCA signals from both {\it in-vivo} and simulated data. Similar to \cite{sreter2018learned}, we utilize convolution layers instead of fully-connected (FC) layers, to exploit the shared spatial information between neighboring image pixels. Our training policy is a two stage process. We start by training the network on simulated data, and then train the resulting network on {\it in-vivo} data. This hybrid policy allows us to improve the network's performance and to achieve a fully-automated network, in which all the regularization parameters are also learned.    
We compare the performance of CORONA with the commonly practiced SVD approach, the iterative RPCA algorithm and an adaptation of the residual network (ResNet), which is considered to be one of the leading deep architectures for a wide variety of problems \cite{he2016deep}. We show that CORONA outperforms all other approaches in terms of image quality and contrast. 

Unfolding, or unrolling an iterative algorithm, was first suggested by Gregor and LeCun \cite{gregor2010learning} to accelerate algorithm convergence. In the context of deep learning, an important question is what type of network architecture to use. Iterative algorithms provide a natural recurrent architecture, designed to solve a specific problem, such as sparse approximations, channel estimation \cite{samuel2017deep} and more. The authors of \cite{gregor2010learning} showed that by considering each iteration of an iterative algorithm as a layer in a deep network and subsequent concatenation of a few such layers it is possible to train such networks to achieve a dramatic improvement in convergence, i.e., to reduce the number of iterations significantly.


In the context of RPCA, a principled way to construct learnable pursuit architectures for structured sparse and robust low rank models was introduced in \cite{sprechmann2015learning}. The proposed networks, derived from the iteration of proximal descent algorithms, were shown to faithfully approximate the solution of RPCA while demonstrating several orders of magnitude speed-up compared to standard optimization algorithms. 
However, this approach is based on a non-convex formulation in which the rank of the low-rank part (or an upper bound on it) is assumed to be known a priori. This poses a network design limitation, as the rank can vary between different applications or even different realizations of the same application, as in CEUS. Thus, for each choice of the rank upper bound, a new network needs to be trained, which can limit its applicability. In contrast, our approach does not require a-priori knowledge of the rank.
Moreover, the use of convolutional layers exploits spatial invariance and facilitates our training process as it reduces the number of learnable parameters dramatically.

The rest of the paper is organized as follows. In \cref{Sec:Arch} we introduce the mathematical formulation of the low-rank and sparse decomposition. \cref{Sec:Met} describes the protocol of the experiments and technical details regarding the realizations of CORONA and ResNet. \cref{Sec:Res} presents {\it in-silico} as well as {\it in-vivo} results of both the iterative algorithm and the proposed deep networks. Finally, we discuss the results, limitations and further research directions in \cref{Sec:Disc}. 

Throughput the paper, $x$ represents a scalar, ${\bf x}$ a vector, ${\bf X}$ a matrix and ${\bf I}_{N\times N}$ is the $N\times N$ identity matrix. The notation $||\cdot||_p$ represents the standard $p$-norm and $||\cdot||_F$ is the Frobenius norm. Subscript $x_l$ denotes the $l$th element of ${\bf x}$ and ${\bf x}_l$ is the $l$th column of ${\bf X}$. Superscript ${\bf x}^{(p)}$ represents ${\bf x}$ at iteration $p$, $T^*$ denotes the adjoint of T, and ${\bf \bar{A}}$ is the complex conjugate of ${\bf A}$.

\section{Deep learning strategy for RPCA in US}\label{Sec:Arch}
\begin{figure*}[!t]%
\centering
\subfigure[Iterative algorithm for sparse and low-rank separation.]{%
\label{fig:FISTA}%
\includegraphics[height=2in]{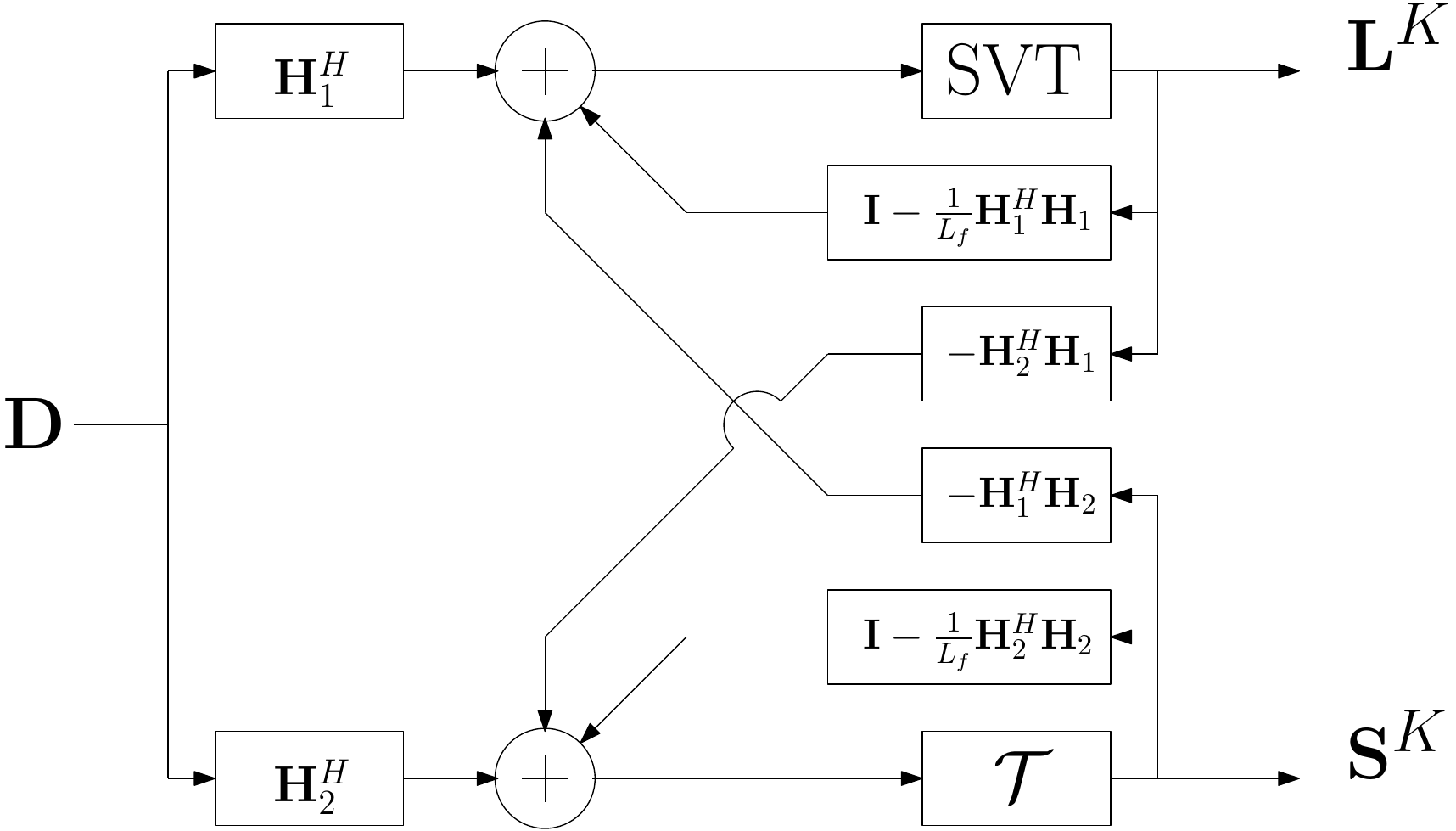}}%
\qquad
\subfigure[Single layer from the corresponding unfolded algorithm.]{%
\label{fig:LISTA}%
\includegraphics[height=2.1in]{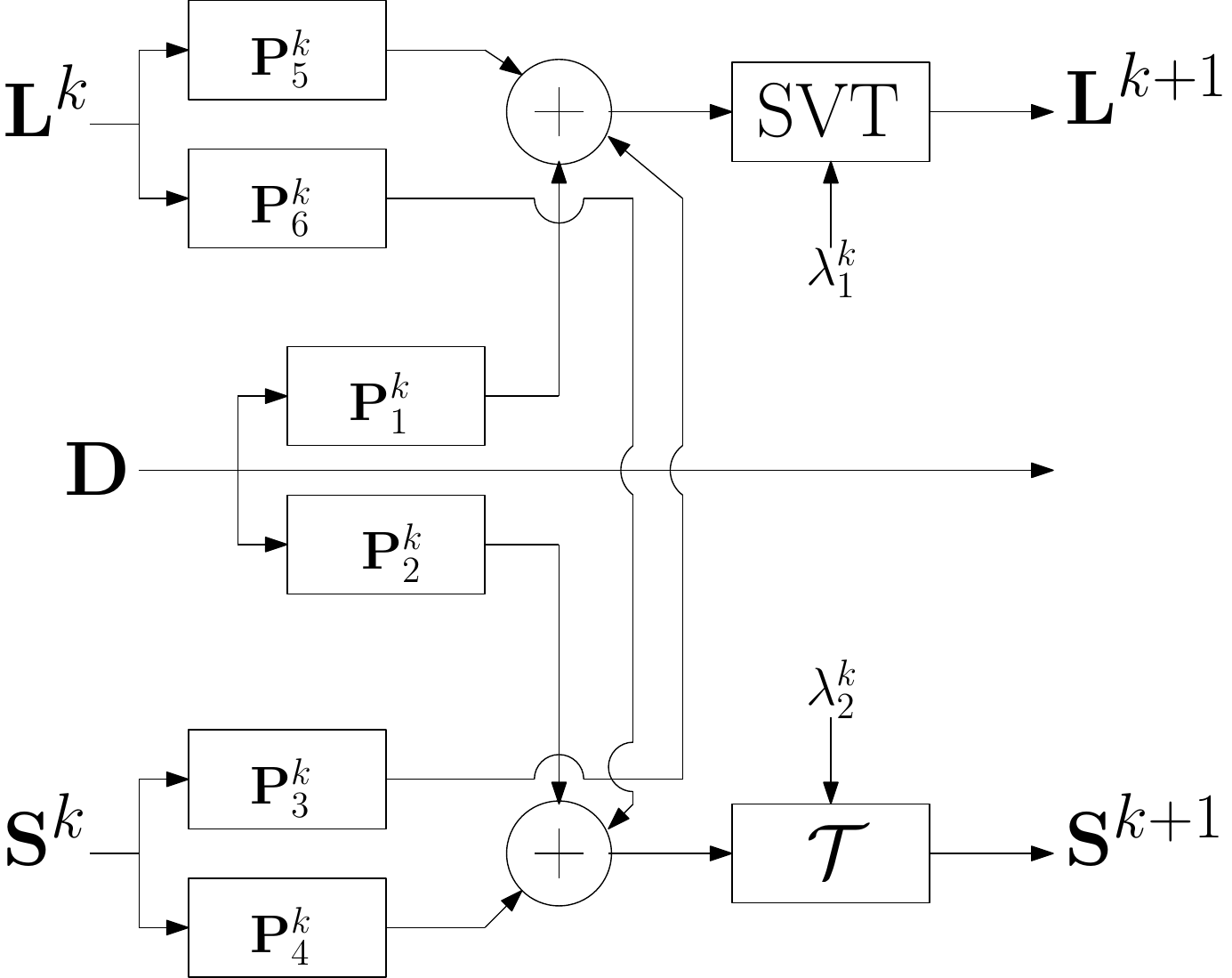}}%
\caption{Architecture comparison between the iterative algorithm applied for $K$ iterations (panel (a)) and its unfolded counterpart (panel (b)). The learned network in panel (b) draws its architecture from the iterative algorithm, and is trained on examples from a given dataset. In both panels, $\v D$ is the input measurement matrix, and $\v S_k$ and $\v L_k$ are the estimated sparse and low-rank matrices in each iteration/layer, respectively.}
\end{figure*}
\subsection{Problem formulation}
We start by providing a low-rank plus sparse (L+S) model for the acquired US signal. In US imaging, typically a series of pulses are transmitted to the imaged medium. The resulting echoes from the medium are received in each transducer element and then combined in a process called beamforming to produce a focused image. As presented in \cite{bar2017fast}, after demodulation the complex analytical (IQ) signal can be represented as 
\begin{equation*}
D(x,z,t)=I(x,z,t)+iQ(x,z,t),
\end{equation*}
where $I(x,z,t)$ and $Q(x,z,t)$ are the in-phase and quadrature components of the demodulated signal, $x,z$ are the vertical and axial coordinates, and $t$ indicates frame number. The signal $D(x,z,t)$ is a sum of echoes returned from the blood/CEUS signal $S(x,z,t)$ as well as from the tissue $L(x,z,t)$, contaminated by additive noise $N(x,z,t)$ 
\begin{equation*}
\label{Eq:LSmodel}
D(x,z,t)=L(x,z,t)+S(x,z,t)+N(x,z,t).
\end{equation*}

Acquiring a series of movie frames $t=1,\ldots,T$, and stacking them as vectors in a matrix $\v D$, leads to the following model
\begin{equation}
\label{Eq:LSmat}
\v D=\v L+\v S + \v N.
\end{equation}
In \cref{Eq:LSmat}, we assume that the tissue matrix $\v L$ can be described as a low-rank matrix, due to its high spatio-temporal coherence. The CEUS echoes matrix $\v S$ is assumed to be sparse, as blood vessels typically sparsely populate the imaged medium. Assuming that each movie frame is of size $M\times M$ pixels, the matrices in \cref{Eq:LSmat} are of size $M^2\times T$. From here on, we consider a more general model, in which the acquired matrix $\v D$ is composed as
\begin{equation}
\label{Eq:LSmat_gen}
\v D=\v H_1\v L+\v H_2\v S + \v N,
\end{equation}
with $\v H_1$ and $\v H_2$ being the measurement matrices of appropriate dimensions.  The model \cref{Eq:LSmat_gen} can also be applied to MR imaging, video compression and additional US applications, as we discuss in \cref{Sec:Disc}. Our goal is to formalize a minimization problem to extract both $\v L$ and $\v S$ from $\v D$ under the corresponding assumptions of L+S matrices.

Similar to \cite{otazo2015low}, we propose solving the following minimization problem
\begin{equation}
\label{Eq:minprob}
\min_{\v L, \v S}\frac{1}{2}||\v D - (\v H_1\v L + \v H_2\v S)||_F^2+\lambda_1||\v L||_{*}+\lambda_2||\v S||_{1,2},
\end{equation}
where $||\cdot||_*$ stands for the nuclear norm, which sums the singular values of $\v L$, and $||\cdot||_{1,2}$ is the mixed $l_{1,2}$ norm, which sums the $l_2$ norms of each row of $\v S$. We use the mixed $l_{1,2}$ norm since the pattern of the sparse outlier (blood or CEUS signal) is the same between different frames, and ultimately corresponds to the locations of the blood vessels, which are assumed to be fixed, or change very slowly during the acquisition period. The nuclear norm is known to promote low-rank solutions, and is the convex relaxation of the non-convex rank minimization constraint \cite{candes2009exact}.  

By defining 
\begin{equation*}
\v X = \left[
\begin{array}{cc}
\v L\\
\v S
\end{array}
\right],\;
\v P_1 = \left[
\begin{array}{cc}
\v I\\
\v 0
\end{array}
\right],\;
\v P_2 = \left[
\begin{array}{cc}
\v 0\\
\v I
\end{array}
\right]
\end{equation*}
and 
$\v A = [\v H_1, \v H_2]$, \cref{Eq:minprob} can be rewritten as 
\begin{equation}
\label{Eq:minprob2}
\min_{\v L, \v S}\frac{1}{2}||\v D - \v A\v X||_F^2+h(\v X),
\end{equation}
where $h(\v X)=\sum_{i=1}^{2}\lambda_i\rho_i(\v P_i\v X)$ with $\rho_1=||\cdot||_*$ and $\rho_2=||\cdot||_{1,2}$. The minimization problem \cref{Eq:minprob2} is a regularized least-squares problem, for which numerous numerical minimization algorithms exist. Specifically, the (fast) iterative shrinkage/thresholding algorithm, (F)ISTA, \cite{eldar2012compressed,palomar2010convex} involves finding the {\it Moreau's proximal} (prox) mapping \cite{moreau1965proximite, Tan2014} of $h$, defined as
\begin{equation}
\label{Eq:proxdef}
\prox_{h}({\v X})=\argmin_{\v U}\left\{h(\v U)+\frac{1}{2}||{\v U}-{\v X}||_F^2 \right\}.
\end{equation}
Plugging the definition of $\v X$ into \cref{Eq:proxdef} yields
\begin{equation*}
\textrm{prox}_{h}(\v X) = \argmin_{\v U_1, \v U_2}\left\{\lambda_1\rho_1(\v U_1)+\frac{1}{2}||\v U_1 - \v L||_F^2\right.
\end{equation*}
\begin{equation*}
\qquad\qquad\qquad\qquad\left.+\lambda_2\rho_2(\v U_2)+\frac{1}{2}||\v U_2 - \v S||_F^2\right\}.
\end{equation*}
Since $\textrm{prox}_{h}(\v X)$ is separable in $\v L$ and $\v S$, it holds that
\begin{equation}
\label{Eq:proxls}
\textrm{prox}_{h}(\v X) =\left[
\begin{array}{cc}
\textrm{prox}_{\rho_1}(\v L)\\
\textrm{prox}_{\rho_2}(\v S)
\end{array}\right]=
\left[
\begin{array}{cc}
\textrm{SVT}_{\lambda_1}(\v L)\\
\Tscr_{\lambda_2}(\v S)
\end{array}\right]
.
\end{equation}
The operators 
$$\Tscr_{\alpha}(\v x)=\max(0,1-\alpha/||\v x||_2)\v x$$ 
and 
$$\textrm{SVT}_{\alpha}(\v X)=\v U\textrm{diag}(\max(0,\sigma_i-\alpha))\v V^H,\;i=1,\ldots,r$$ 
are the mixed $l_{1/2}$ soft thresholding \cite{eldar2012compressed} and singular value thresholding \cite{cai2010singular} operators. 
Here $\v X$ is assumed to have an SVD given by $\v X=\v U\v \Sigma\v V^H$ with $\v \Sigma=\textrm{diag}(\sigma_i,\ldots,\sigma_r)$, a diagonal matrix of the eigenvalues of $\v X$.
The proximal mapping \cref{Eq:proxls} is applied in each iteration to the gradient of the quadratic part of \cref{Eq:minprob2}, given by
\begin{equation*}
g(\v X)=\frac{d}{d\v X}\frac{1}{2}||\v D - \v A\v X||_F^2 = \v A^H(\v A\v X - \v D),
\end{equation*}
and more specifically,
\begin{equation*}
\left[
\begin{array}{cc}
\frac{d}{d\v L}\\
\frac{d}{d\v S}
\end{array}
\right]=
\left[
\begin{array}{cc}
\v H_1^H(\v H_1\v L + \v H_2\v S - \v D)\\
\v H_2^H(\v H_1\v L + \v H_2\v S - \v D)
\end{array}
\right].
\end{equation*}

The general iterative step of ISTA applied to minimizing \cref{Eq:minprob} (L+S ISTA) is thus given by 
\begin{equation*}
\v X^{k+1}=\textrm{prox}_h\left(\v X^k-\frac{1}{L_f}g(\v X^k)\right),
\end{equation*}
or
\footnotesize
\begin{equation}
\label{Eq:Iterative}
\begin{array}{ll}
\v L^{k+1} = \textrm{SVT}_{\lambda_1/L_f}\left\{\left(\v I-\frac{1}{L_f}\v H_1^H\v H_1\right)\v L^{k} - \v H_1^H\v H_2\v S^{k} + \v H_1^H\v D\right\} \\
\v S^{k+1} = \Tscr_{\lambda_2/L_f}\left\{\left(\v I-\frac{1}{L_f}\v H_2^H\v H_2\right)\v S^{k} - \v H_2^H\v H_1\v L^{k} + \v H_2^H\v D\right\}
\end{array},
\end{equation}\normalsize
where $L_f$ is the Lipschitz constant of the quadratic term of \cref{Eq:minprob2}, given by the spectral norm of $\v A^H\v A$. 

The L+S ISTA algorithm for minimizing \cref{Eq:minprob} is summarized in \cref{Alg:ISTA}. The diagram in \cref{fig:FISTA} presents the iterative algorithm, which relies on knowledge of $\v H_1, \v H_2$ and selection of $\lambda_1$ and $\lambda_2$. 

    \begin{algorithm}
        \caption{L+S ISTA for minimizing \cref{Eq:minprob}}
        \label{Alg:ISTA}
        \begin{algorithmic}
            \REQUIRE $\v D$, $\lambda_1>0, \lambda_2>0$, maximum iterations $K_{\textrm{max}}$
            \STATE {\bf Initialize} $\v S = \v L = \v 0$ and $k=1$
            \WHILE {$k\leq K_{\textrm{max}}$ or stopping criteria not fulfilled}
            	\STATE {\bf 1:} $\v G_{1_k}=\left(\v I-\frac{1}{L_f}\v H_1^H\v H_1\right)\v L^{k} - \v H_1^H\v H_2\v S^{k} + \v H_1^H\v D$
                \STATE {\bf 2:} $\v G_{2_k} = \left(\v I-\frac{1}{L_f}\v H_2^H\v H_2\right)\v S^{k} - \v H_2^H\v H_1\v L^{k} + \v H_2^H\v D$
                \STATE {\bf 3:} $\v L^{k+1} = \textrm{SVT}_{\lambda_1/L_f}\left\{\v G_{1_k}\right\}$
                \STATE {\bf 4:} $\v S^{k+1} = \Tscr_{\lambda_2/L_f}\left\{\v G_{2_k}\right\}$
                \STATE {\bf 5:} $k\leftarrow k+1$
            \ENDWHILE
            \RETURN $\v S_{K_{\textrm{max}}}, \v L_{K_{\textrm{max}}}$
        \end{algorithmic}
    \end{algorithm}

The dynamic range between returned echoes from the tissue and UCA/blood signal can range from $10$dB to $60$dB. As this dynamic range expands, more iterations are required to achieve good separation of the signals. This observation motivates the pursuit of a fixed complexity algorithm. In the next section we propose CORONA which is based on unfolding \cref{Alg:ISTA}. Background on learning fast sparse approximations is given in \cref{Sec:ISTA} of the supplementary materials. 

\subsection{Unfolding the iterative algorithm}
An iterative algorithm can be considered as a recurrent neural network, in which the $k$th iteration is regarded as the $k$th layer in a feedforward network \cite{sprechmann2015learning}. To form a convolutional network, one may consider convolutional layers instead of matrix multiplications. With this philosophy, we form a network from \cref{Eq:Iterative} by replacing each of the matrices dependent on $\v H_1$ and $\v H_2$ with convolution layers (kernels) $\v P_1^k,\ldots,\v P_6^k$ of appropriate sizes. These will be learned from training data. Contrary to previous works in unfolding RPCA which considered training fully connected (FC) layers \cite{sprechmann2015learning}, we employ convolution kernels in our implementation which allows us to achieve spatial invariance while reducing the number of learned parameters considerably. 

The kernels as well as the regularization parameters $\lambda_1^k$ and $\lambda_2^k$ are learned during training. By doing so, the following equations for the $k$th layer are obtained
\begin{align*}
&\v L^{k+1} = \textrm{SVT}_{\lambda_1^k}\left\{\v P_5^k*\v L^{k} + \v P_3^k*\v S^{k} + \v P_1^k*\v D\right\}, \\
&\v S^{k+1} = \Tscr_{\lambda_2^k}\left\{\v P_6^k*\v L^{k} + \v P_4^k*\v S^{k} + \v P_2^k*\v D\right\},
\end{align*}
with $*$ being a convolution operator. The latter can be considered as a single layer in a multi-layer feedforward network, which we refer to as CORONA: Convolutional rObust pRincipal cOmpoNent Analysis. A diagram of a single layer from the unfolded architecture is given in \cref{fig:LISTA}, where the supposedly known model matrices were replaced by the 2D convolution kernels $\v P_1^k,\ldots,\v P_6^k$, which are learned as part of the training process of the overall network. 

In many applications, the recovered matrices $\v S$ and $\v L$ represent a 3D volume, or movie, of dynamic objects imposed on a (quasi) static background. Each column in $\v S$ and $\v L$ is a vectorized frame from the recovered sparse and low-rank movies. Thus, we consider in practice our data as a 3D volume and apply 2D convolutions. The SVT operation (which has similar complexity as the SVD operation) at the $k$th layer is performed after reshaping the input 3D volume into a 2D matrix, by vectorizing and column-wise stacking each frame. 

The thresholding coefficients are learned independently for each layer. Given the $k$th layer, the actual thresholding values for both the SVT and soft-thresholding operations are given by
$
\textrm{thr}_L^k=\sigma(\lambda_L^k)\cdot a_L\cdot\max(L^k)
$
and
$
\textrm{thr}_S^k=\sigma(\lambda_S^k)\cdot a_S\cdot\textrm{mean}(S^k)
$
respectively, where $\sigma(x)=1/(1+\exp(-x))$ is a sigmoid function, $a_L$ and $a_S$ are fixed scalars (in our application we chose $a_L=0.4$ and $a_S=1.8$) and $\lambda_L^k$ and $\lambda_S^k$ are learned in each layer by the network. 

\begin{figure*}[ht] 
    \centering
    \includegraphics[width=1\linewidth]{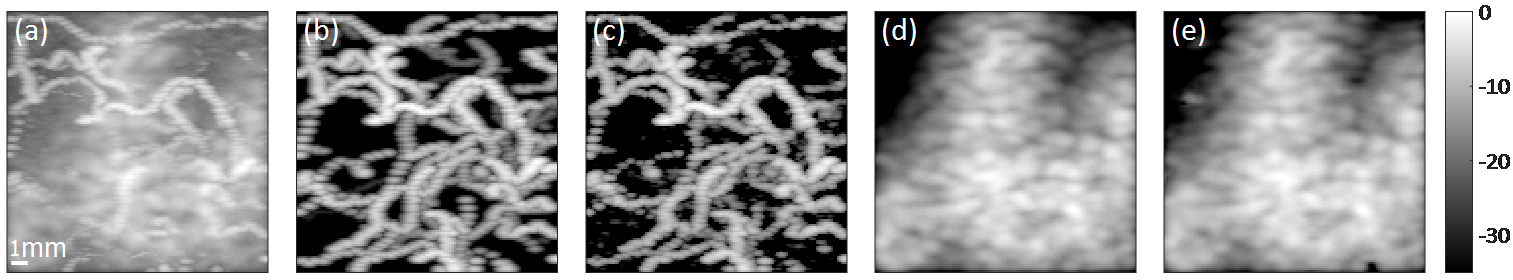}
    \caption{Simulation results of CORONA. (a) MIP image of the input movie, composed from 50 frames of simulated UCAs cluttered by tissue. (b) Ground-truth UCA MIP image. (c) Recovered UCA MIP image via CORONA. (d) Ground-truth tissue MIP image. (d) Recovered tissue MIP image via CORONA. Color bar is in dB.}\label{fig:sim_unfolded}
\end{figure*} 


\subsection{Training CORONA}\label{Sec:TrainingCORONA}
CORONA is trained using back-propagation in a supervised manner. Generally speaking, we obtain training examples $\v D_i$ and corresponding sparse $\hat{\v S_i}$ and low-rank $\hat{\v L_i}$ decompositions. In practice, $\hat{\v S_i}$ and $\hat{\v L_i}$ can either be obtained from simulations or by decomposing $\v D_i$ using iterative algorithms such as FISTA \cite{Beck2009}. 
The loss function is chosen as the sum of the mean squared errors (MSE) between the predicted $\v S$ and $\v L$ values of the network and $\hat{\v S_i}$, $\hat{\v L_i}$, respectively, 
\begin{dmath}
\label{Eq:mse_loss_function}
\mathcal{L}({\bm \theta}) = \frac{1}{2N}\sum_{i=1}^{N}||f_{S}(\v D_i, {\bm \theta}) - \hat{\v S}_i||_F^2\\+\frac{1}{2N}\sum_{i=1}^{N}||f_{L}(\v D_i, {\bm \theta}) - \hat{\v L}_i||_F^2.
\end{dmath}
In the latter equation, $f_{S}/f_{L}$ is the sparse/low-rank output of CORONA with learnable parameters ${\bm \theta}=\{\v P_1^k,\ldots,\v P_6^k,\lambda_1^k,\lambda_2^k\}$, $k=1,\ldots,K$, where $K$ is the number of chosen layers. 

Training a deep network typically requires a large amount of training examples, and in practice, US scans of specific organs are not available in abundance. To be able to train CORONA, we thus rely on two strategies: patch-based analysis and simulations. Instead of training the network over entire scans, we divide the US movie used for training into 3D patches (axial coordinate, lateral coordinate and frame number). Then we apply \cref{Alg:ISTA} on each of these 3D patches. The SVD operations in \cref{Alg:ISTA} become computationally tractable since we work on relatively small patches. The resulting separated UCA movie is then considered as the desirable outcome of the network and the network is trained over these pairs of extracted 3D patches from the acquired movie, and the resulting reconstructed UCA movies. In practice, the CEUS movie used for training is divided into 3D patches of size $32\times 32\times 20$ ($32\times 32$ pixels over 20 consecutive frames) with 50\% overlap between neighboring patches. The regularization parameters of \cref{Alg:ISTA}, $\lambda_1$ and $\lambda_2$ are chosen empirically, but are chosen once for all the extracted patches. 

In \cref{Sec:Sims_SI} of the supplementary materials, we provide a detailed description of how the simulations of the UCA and tissue movies were generated. In particular, we detail how individual UCAs were modeled and propagated in the imaging plane, and describe the cluttering tissue signal model. We then demonstrate the importance of training on both simulations and {\it in-vivo} data in \cref{Sec:SI_TraningImportance} of the supplementary materials.




\section{Experiments}\label{Sec:Met}
The brains of two rats were scanned using a Vantage 256 system (Verasonics Inc., Kirkland, WA, USA). An L20-10 probe was utilized, with a central frequency of 15MHz. The rats underwent craniotomy after anesthesia to obtain an imaging window of $6\times2\textrm{mm}^2$. A bolus of $100\mu$L $\textrm{SonoVue}^{\textrm{TM}}$ (Bracco, Milan, Italy) contrast agent, diluted with normal saline with a ratio of 1:4, was administered intravenously to the rat’s tail vein. Plane-wave compounding of five steering angles (from $-12^\circ$ to $12^\circ$, with a step of $6^\circ$) was adopted for ultrasound imaging. For each rat, over $6000$ consecutive frames were acquired with a frame rate of 100Hz. 300 frames with relatively high B-mode intensity were manually selected for data processing in this work.

In recent years, several deep learning based architectures have been proposed and applied successfully to classification problems. One such approach is the residual network, or ResNet \cite{he2016deep}. ResNet utilizes convolution layers, along with batch normalization and skip connections, which allow the network to avoid vanishing gradients and reduce the overall number of network parameters. 

To compare with CORONA, we implemented ResNet using complex convolutions for the tissue clutter suppression task. The network does not recover the tissue signal, as CORONA, but only the UCA signal. In \cref{Sec:Res} and in the supporting materials file, we compare both architectures and assess the advantages and disadvantages of each network. 
In \cref{Sec:InVivo}, we show that CORONA outperforms ResNet in terms of image quality (contrast) of the CEUS signal.

Both ResNet and CORONA were implemented in Python 3.5.2, using the PyTorch 0.4.1 package. CORONA consists of 10 layers. First three layers used convolution filters of size $5\times 5\times 1$ with stride $(1,1,1)$, padding $(2,2,0)$ and bias, while the last seven layers used filters of size $3\times 3\times 1$ with stride $(1,1,1)$, padding $(1,1,0)$ and bias. Training was performed using the ADAM optimizer with a learning rate of 0.002. For the {\it in-vivo} experiments in \cref{Sec:Res}, we trained the network over $2400$ simulated training pairs and additional $2400$ {\it in-vivo} pairs taken only from the first rat. Training pairs were generated from the acquired US clips, after dividing each clip to $32\times32\times20$ patches. We then applied \cref{Alg:ISTA} for each patch with $\lambda_1=0.02$, $\lambda_2=0.001$ and $D_{\textrm{max}}=30000$ iterations to obtain the separated UCA signal for the training process. \cref{Alg:ISTA} was implemented in MATLAB (Mathworks Inc.) and was applied to the complex-valued IQ signal. PyTorch performs automatic differentiation and back-propagation using the Autograd functionality, and version 0.4.1 also supports back-propagation through SVD, but only for real valued numbers. Thus, complex valued convolution layers and SVD operations were implemented. 

\section{Results}\label{Sec:Res}
\begin{figure}[!t] 
	\vspace{-0.7cm}
    \centering
    \includegraphics[width=1\linewidth]{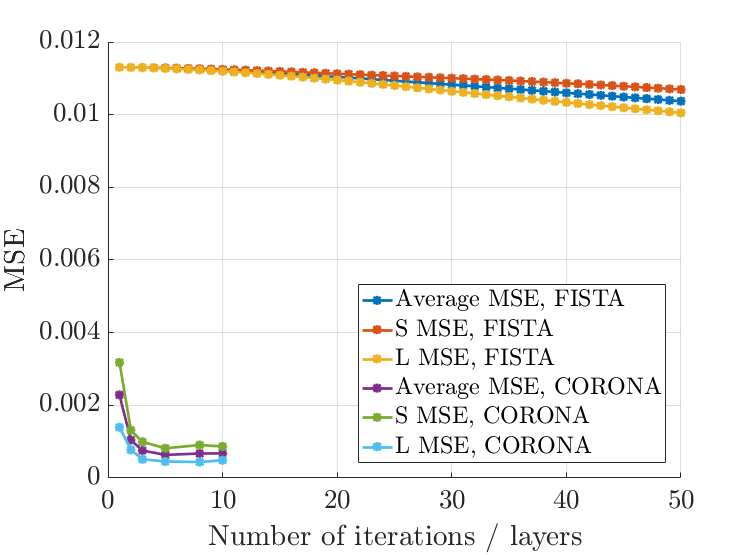}
    \caption{MSE plot for the FISTA algorithm and CORONA as a function of the number of iterations/layers.}\label{Fig:mse}
\end{figure}

\begin{figure*}[!ht] 
    \centering
    \includegraphics[width=0.9\linewidth]{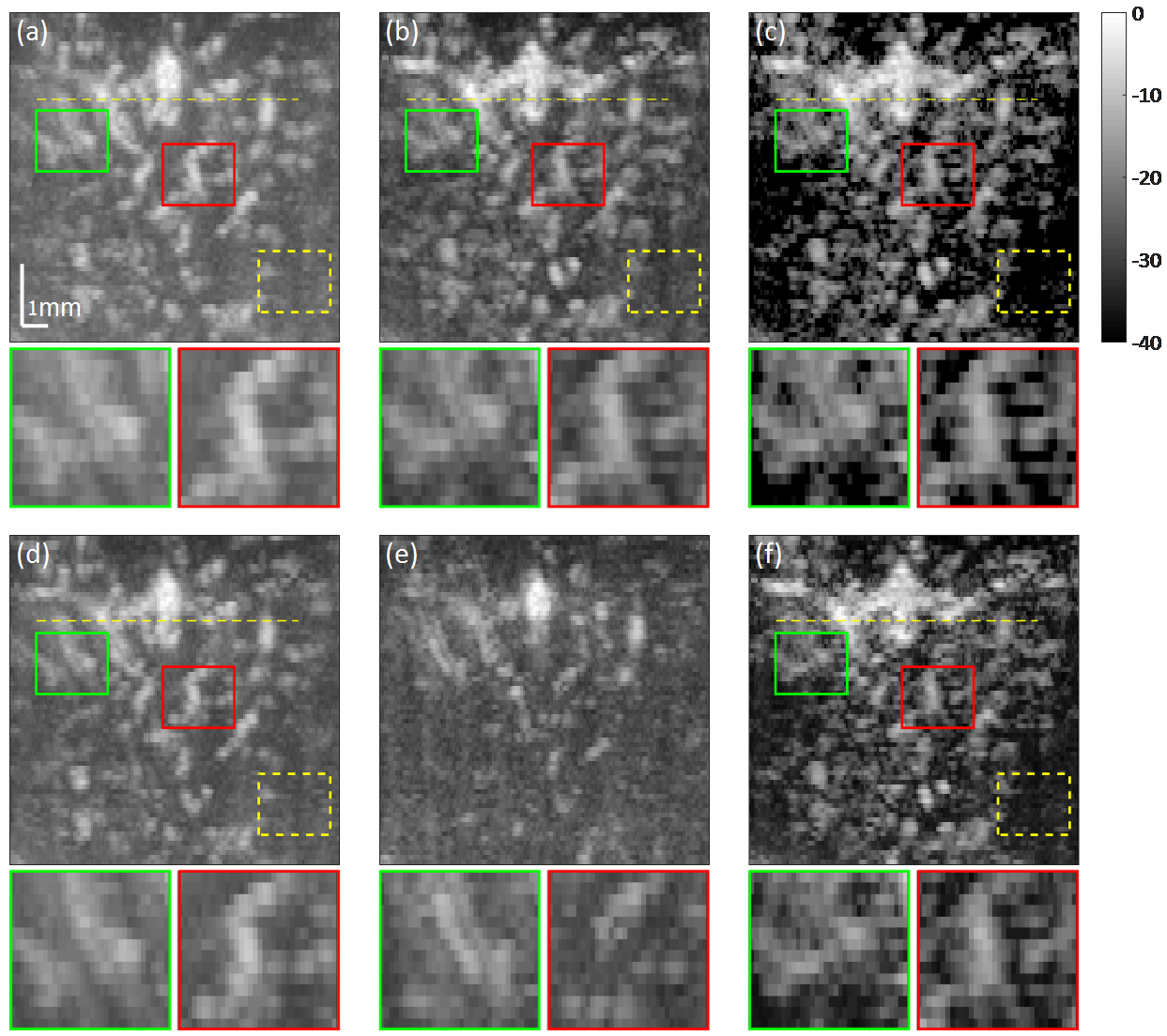}
    \caption{Recovery of {\it in-vivo} CEUS signal depicting rat brain vasculature. (a) SVD based separation. (b) L+S FISTA separation. (c) Deep network separation, with the unfolded architecture of the FISTA algorithm. (d) Wall filtering with cutoff frequency of $0.2\pi$ (e) Wall filtering with cutoff frequency of $0.9\pi$ (f) ResNet. Color bar is in dB.}\label{Fig:main}
\end{figure*}  

\subsection{Simulation results}
In this section we provide reconstruction results for CORONA applied to a simulated dataset, and trained on simulations. Figure \ref{fig:sim_unfolded} presents reconstruction results of the UCA signal $\v S$ and the low-rank tissue $\v L$ against the ground truth images. Panel (a) shows a representative image in the form of maximum intensity projection (MIP)\footnote{In order to present a single representative image, we take the pixel-wise maximum from each movie. This process is also referred to as maximum intensity projection, and is a common method to visualize CEUS images.} of the input cluttered movie (50 frames). It is evident that the UCA signal, depicted as randomly twisting lines, is masked considerably by the simulated tissue signal. Panel (b) illustrates the ground truth MIP image of the UCA signal, while panel (c) presents the MIP image of the recovered UCA signal via CORONA. Panels (d) and (e) show MIP images of the ground truth and CORONA recovery, respectively. 

Observing all panels, it is clear that CORONA is able to recover reliably both the UCA signal and the tissue signal. 
\cref{Sec:SI_re} in the supporting materials provides additional simulation results, showing also the recovered UCA signal by ResNet. Although qualitatively ResNet manages to recover well the UCA signal, its contrast is lower than the contrast of the CORONA recovery, which presents a clearer depiction of the random vascular structure of the simulation. Moreover, ResNet does not recover the tissue signal, while CORONA does. 



As CORONA draws its architecture from the iterative ISTA algorithm, our second aim in this section is to assess the performance of both CORONA and the FISTA algorithm by calculating the MSE of each method as a function of iteration/layer number. Each layer in CORONA can be thought of as an iteration in the iterative algorithm. 
To that end, we next quantify the MSE over the simulated validation batch (sequence of 100 frames) as a function of layer number (CORONA) and iteration number (FISTA), as presented in \cref{Fig:mse}. For both methods, the MSE for the recovered sparse part (UCA signal) $\v S$ and the low-rank part (tissue signal) $\v L$ were calculated as a function of iteration/layer number, as well as the average MSE of both parts, according to (\ref{Eq:mse_loss_function}) ($\alpha=0.5$). For each layer number, we constructed an unfolded network with that number of layers, and trained it for 50 epochs on simulated data only.

Observing \cref{Fig:mse}, it is clear that even when considering CORONA with only 1 layer, its performance in terms of MSE in an order of magnitude better than FISTA applied with 50 iterations. Adding more layers improves the CORONA MSE, though after 5 layers, the performance remains roughly the same. Figure \ref{Fig:mse} also shows that a clear decreasing trend is present for the FISTA MSE, however a dramatic increase in the number of iterations is required by FISTA to achieve the same MSE values. 

\subsection{In-vivo experiments}\label{Sec:InVivo}
We now proceed to demonstrate the performance of CORONA on {\it in-vivo} data. As was described in \cref{Sec:Met}, CORONA was trained on both simulated and experimental data. In \cref{Fig:main}, panel (a) depicts SVD based separation of the CEUS signal, panel (b) shows the FISTA based separation and panel (c) shows the result of CORONA. 
The lower panels of \cref{Fig:main} also compare the performance of the trained ResNet (panel (f)) on the {\it in-vivo} data as well as provide additional comparison to the commonly used wall filtering. Specifically, we use a $6$th order Butterworth filter with two cutoff frequencies of $0.2\pi$ (panel (d)) and $0.9\pi$ (panel (e)) radians/samples. Two frequencies were chosen which represent two scenarios. The cutoff frequency of the recovery in panel (d) was chosen to suppress as much tissue signal as possible, without rejecting slow moving UCAs. In panel (e), a higher frequency was chosen, to suppress the slow moving tissue signal even further, but as can be seen, at a cost of removing also some of the slower bubbles. The result is a less consistent vascular image.  
Visually judging, all panels of \cref{Fig:main} shows that ResNet outperforms both the SVD and wall filtering approaches. However, a more careful observation shows that the ResNet output, although more similar to CORONA's output, seems more grainy and less smooth than CORONA's image. CORONA's recovery exhibits the highest contrast, and produces the best visual.

In each panel, the green and red boxes indicate selected areas, whose enlarged views are presented in the corresponding green and red boxes below each panel. Visual inspection of the panels (a)-(f) shows that FISTA, ResNet and CORONA achieve CEUS signal separation which is less noisy than the naive SVD approach and wall filtering. Considering the enlarged regions below the panels further supports this conclusion, showing better contrast of the FISTA and deep networks outputs. The enlarged panels below panels (d) and (e) show that indeed, as the cutoff frequency of the wall filter is increased, slow moving UCAs are also filtered out. Both deep networks exhibit higher contrast than the other approaches. 

\begin{table}[h!]
  \begin{center}
    \caption{\small CNR values for the selected green and red rectangles of \cref{Fig:main}, as compared with the dashed yellow background rectangle in each corresponding panel. All values are in dB.} 
    \label{tab:table1}
    \footnotesize 
    \begin{tabular}{|c|c|c|c|c|s|}
      \hline 
      & \textbf{SVD} & \textbf{Wall filter} & \textbf{FISTA} & \textbf{ResNet} & \textbf{Unfolded} \\
      \hline 
      Green box & -1.65 & -2.02 & -1.67 & -2.17 & -0.3 \\
      Red box   & -4.8  & -5.55 & -3.52 & -2.95 & -1.13\\
      \hline 
    \end{tabular}
  \end{center}
\end{table}

\begin{table}[h!]
  \begin{center}
    \caption{\small CR values for the selected green and red rectangles of \cref{Fig:main}, as compared with the dashed yellow background rectangle in each corresponding panel. All values are in dB.} 
    \label{tab:table2}
    \footnotesize
    \begin{tabular}{|c|c|c|c|c|s|}
      \hline 
      & \textbf{SVD} & \textbf{Wall filter} & \textbf{FISTA} & \textbf{ResNet} & \textbf{Unfolded}\\
      \hline 
      Green box & 4.68 & 4.5 & 5.52 & 7.92 & 15.24\\
      Red box   & 4.56 & 4.1 & 5.24 & 7.55 & 14.88 \\
      \hline 
    \end{tabular}
  \end{center}
\end{table}

To further quantify the performance of each method, we provide two metrics to assess the contrast ratio of their outputs, termed contrast to noise ratio (CNR) and contrast ratio (CR). 

CNR is calculated between a selected patch, e.g. the red or green boxes in panels (a)-(f) and a reference patch, marked by the dashed yellow patches, which represents the background, for the same image. That is, for each panel we estimate the CNR values of the red - yellow and green - yellow boxes, where $\mu_s$ is the mean of the red/green box with variance $\sigma_{s}^{2}$ and $\mu_b$ is the mean of the dashed yellow patch with variance $\sigma_{b}^{2}$. The CNR is defined as
\begin{equation*}\label{Eq:CNR}
\textrm{CNR}=\frac{|\mu_s-\mu_b|}{\sqrt{\sigma_{s}^{2}+\sigma_{b}^{2}}}.
\end{equation*}
Similarly, the CR is defined as
\begin{equation*}\label{Eq:CNR}
\textrm{CR}=\frac{\mu_s}{\mu_b}.
\end{equation*}
\cref{tab:table1} and \cref{tab:table2} provide the calculated CNR and CR values of each method, respectively. 

\begin{figure}[!t] 
    \centering
    \includegraphics[width=0.8\linewidth]{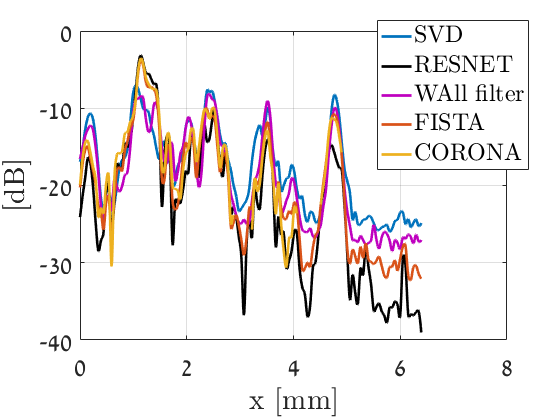}
    \caption{Intensity profiles across the dashed yellow lines in panels (a)-(c) of \cref{Fig:main}. Regions in which CORONAs' curve is missing indicate a value of $-\infty$. Values are in dB.}\label{Fig:intensity}
\end{figure}  


In both metrics, higher values imply higher contrast ratios, which suggest better noise suppression and better signal depiction. Considering both tables, CORONA outperforms all other approaches. In most cases, its performance is an order of magnitude better than that SVD. The CR values of ResNet are also better than the baseline SVD, though lower than those of CORONA. Its CNR values however, are not always higher than those of the SVD. In terms of CR, the FISTA results are better than those of the SVD filter, though lower than the deep-learning based approaches. In terms of CNR, for the green box, FISTA is comparable to SVD and better than ResNet, while for the red box, its performance is the worst. 

Both metrics support the previous conclusions, that by combining a proper model to the separation problem with a data-driven approach leads to improved separation of UCA and tissue signals, as well as noise reduction as compared to the popular SVD approach.  

Finally, we also provide intensity cross-sections, taken along the horizontal yellow dashed line for each method, as presented in \cref{Fig:intensity}. Considering the intensity cross-section of \cref{Fig:intensity}, it is evident that all methods reconstruct the peaks with good correspondence. The FISTA and deep-learning networks' profiles exhibit higher contrast than the SVD and wall filter (deeper ``cavities"). In some areas, the unfolded (yellow) profiles seems to vanish. This is because the attained value is $-\infty$. The supporting materials file contains additional comparisons. \cref{Sec:SI_losses} presents the training and validation losses of the networks, as well as the evolution of the regularization coefficients of CORONA as a function of epoch number. \cref{Sec:SI_TraningImportance} discusses the importance of training the networks on both simulations and {\it in-vivo data} when applying CORONA on {\it in-vivo} experiments, while \cref{Sec:SI_runtime} presents the training and execution times for both networks. 

\section{Discussion and conclusions}\label{Sec:Disc}
In this work, we proposed a low-rank plus sparse model for tissue/UCA signal separation, which exploits both spatio-temporal relations in the data, as well as the sparse nature of the UCA signal. This model leads to a solution in the form of an iterative algorithm, which outperforms the commonly practiced SVD approach. We further suggested to improve both execution time and reconstructed image quality by unfolding the iterative algorithm into a deep network, referred to as CORONA. The proposed architecture utilizes convolution layers instead of FC layers and a hybrid simulation-{\it in-vivo} training policy. Combined, these techniques allow CORONA to achieve improved performance over its iterative counterpart, as well as over other popular architectures, such as ResNet. 
We demonstrated the performance of all methods on both simulated and {\it in-vivo} datasets, showing improved vascular depiction in a rat's brain. 

We conclude by discussing several points, regarding the performance and design of deep-learning based networks.   
First, we attribute the improved performance over the commonly practiced SVD filtering, wall filtering and FISTA to two main reasons. The first, is the fact that for application on {\it in-vivo} data, the networks are trained based on both {\it in-vivo} data and simulated data. The simulated data provides the networks with an opportunity to learn from ``perfect" examples, without noise and with absolute separation of UCAs and their surroundings. In \cref{Sec:SI_TraningImportance} of the supplementary materials we show the effect on recovery when the network is trained with and without experimental data.  
The iterative algorithm, on the other hand, cannot learn or improve its performance on the {\it in-vivo} data from the simulated data. The second, is the fact that both networks rely on 2D complex convolutions. Contrary to FC layers, convolution layers reduce the number of learnable parameters considerably, thus help avoid over-fitting and achieve good performance even when the training sets are relatively low. Moreover, convolutions offer spatial invariance, which allows the network to capture spatially translated UCAs. 

Focusing on patch-based training (\cref{Sec:TrainingCORONA}) over entire image training has several benefits. UCAs are used to image blood vessels, and as such entire images will include implicitly blood vessel structure. Thus, training over entire images may result in the network being biased towards the vessel trees presented in the (relatively small) training cohort. On the other hand, small patches are less likely to include meaningful structure, hence training on small patches will be less likely to bias the network towards specific blood vessel structures and enable the network to generalize better. Furthermore, as FISTA and CORONA employ SVD operations, processing the data in small batches improves execution time \cite{song2017accelerated,song2017ultrasound}.

Second, as was mentioned in the introduction, in the context of RPCA, a principled way to construct learnable pursuit architectures for structured sparse and robust low rank models was introduced in \cite{sprechmann2015learning}. The proposed network was shown to faithfully approximate the RPCA solution with several orders of magnitude speed-up compared to its standard optimization algorithm counterpart. 
However, this approach is based on a non-convex formulation of the nuclear norm in which the rank (or an upper bound of it) is assumed to be known a priori. 

The main idea in \cite{sprechmann2015learning} is to majorize the non-differentiable nuclear norm with a differentiable term, such that the low-rank matrix is factorized as a product of two matrices, $\v L=\v A\v B$, where $\v A\in \bbR^{n\times q}$ and $\v B\in \bbR^{q\times m}$. Using this kind of factorization alleviates the need to compute the SVD product, but introduces another unknown parameter $q$ which needs to be set (typically by hand), and corresponds to the rank of the low-rank matrix. This poses a network design limitation, as the rank can vary between different applications or even different realizations of the same application, requiring the network to be re-trained per each new choice of $q$.

In fact, this is the same rank-thresholding parameter as in the standard SVD filtering technique, which we want to avoid hand-tuning. 
Moreover, this kind of factorization leads to a non-convex minimization problem, whose globally optimal stationary points depend on the choice of the regularization parameter $\lambda_*$. Since typically these parameters are chosen empirically, a wrong choice of $\lambda_*$ may lead to suboptimal reconstruction results of the RPCA problem, which are then used as training data for the fixed complexity learned algorithm. 
Since we operate on the original convex problem, we train against optimal reconstruction results of the RPCA algorithm, without the need to a-priori estimate the low-rank degree, $q$. 

Third, currently CORONA and ResNet offer a trade-off between them. By relying on convolutions, CORONA is trained with a considerable lower number of parameters (314 for 1 layer, 1796 for 10 layers) than the ResNet (25378). CORONA outperforms ResNet in both visual quality and quantifiable metrics, as presented in \cref{Sec:Res}. However, its training and execution times are slower (see \cref{Sec:SI_runtime} in the supporting materials file). This performance-runtime trade-off is attributed to the fact that CORONA relies on SVD decomposition in each layer, which is a relatively computationally demanding operation.  However, it allows the network to learn the rank of the low-rank matrix, without the need to upper bound it and restrict the architecture of the network. Incorporation of fast approximations for SVD computations, 
such as truncated or random SVD \cite{wang2018fast,song2017accelerated,halko2011finding,martinsson2016randomized}, can potentially expedite the network's performance and achieve faster execution than ResNet. 
It is also important to keep in mind that ResNet does not recover the tissue signal, only the UCA signal. In some applications, such as super-resolution CEUS imaging over long time durations, the tissue signal is used to correct for motion artifacts.

On a final note, the proposed iterative and deep methods were demonstrated on the extraction of CEUS signal from an acquired IQ movie, but in principle can also be applied to dynamic MRI sequences, as well as to the separation of blood from tissue, e.g. for Doppler processing. In the latter case, the dynamic range between the tissue signal and the blood signal will be greater than that of the tissue and UCA signal. In terms of the iterative algorithm, this would lead to more iterations for the separation process, but once the iterative algorithm has finished, its learned version could be trained on its output to achieve faster execution. 



\section*{Acknowledgment}
The authors would like to thank De Ma and Zhifei Dai from the Biomedical Engineering department of Peking university for help in performing the {\it in-vivo} experiments.

\def\IEEEbibitemsep{0pt plus .5pt}
\small
\bibliographystyle{lib/myIEEEtran}
\bibliography{Bib_Mendeley}

\newpage
\setcounter{section}{0}
\onecolumn
\doublespacing
\large 

\begin{figure}[h]
\centering
\huge{
Deep Unfolded Robust PCA with Application to \\ Clutter Suppression in Ultrasound \\ supporting materials}
\end{figure}
\section{Learning fast approximations via unfolding}
\label{Sec:ISTA}
To better understand the concept of unfolding an iterative algorithm, we briefly describe the basic ideas presented in \cite{gregor2010learning}. Consider the following sparse recovery model
\begin{equation*}
\v y = \v A\v x,
\end{equation*}
\begin{figure}[ht] 
    \centering
    \includegraphics[width=.7\linewidth]{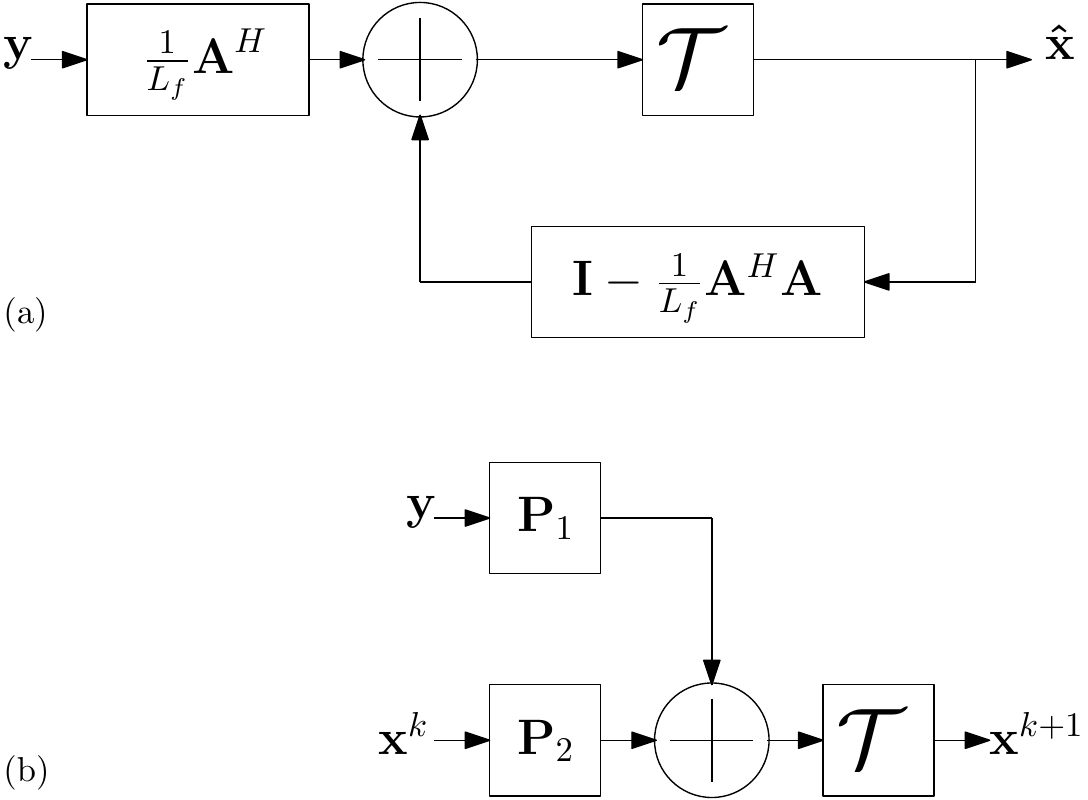}
    \caption{ISTA iterative algorithm (panel (a)) compared with the learned ISTA (panel (b)). Each iteration in the iterative algorithm is replaced with a single layer in the learned algorithm. Instead of using the model parameters such as $\v A$, these parameters are replaced with general matrices $\v P_1$ and $\v P_2$ which are learned.}\label{fig:ista_r}
\end{figure}  
where $\v y$ is a length-$m$ measurement vector, $\v x$ is a length-$n$ sparse vector to be recovered, and $\v A$ is the sensing matrix. Recovering $\v x$ from $\v y$ can be performed by formulating the following convex minimization problem
\begin{equation}
\label{Eq:sparserec}
\min_{\v x}\frac{1}{2}||\v y-\v A\v x||_2^2+\lambda||\v x||_1,
\end{equation}
where $\lambda>0$ is a regularization parameter. A popular iterative algorithm which minimizes \cref{Eq:sparserec} is the ISTA algorithm, or its faster counterpart, the fast ISTA (FISTA). FISTA is guaranteed to converge, in the worst case scenario, with a rate proportional to $1/k^2$, with $k$ being the iteration number. As suggested in \cite{gregor2010learning}, this convergence can be sped up by proposing a learned version of ISTA (LISTA). Furthermore, the authors of \cite{giryes2018tradeoffs} demonstrated that the unfolded architecture facilitates a trade-off between fast convergence and reconstruction accuracy of the sparse recovery problem. 

More specifically, the iterative scheme of ISTA consists of the following iterative step
\vspace{0.2cm}
\begin{equation*}
\v x^{k+1}=\Tscr_{\lambda/L_f}\left\{\left(\v I-\frac{1}{L_f}\v A^H\v A\right)\v x^{k}+\frac{1}{L_f}\v A^H\v y\right\},
\end{equation*}
\vspace{0.3cm}
with $\Tscr_{\lambda/L_f}(\cdot)$ being the element-wise soft-thresholding operator with parameter $\lambda/L_f$ and $L_f$ is the spectral norm of $\v A^H\v A$. This iterative procedure is illustrated in panel (a) of \cref{fig:ista_r}, where $\hat{\v x}$ is the output of the ISTA algorithm.

Conversely, we can consider each iteration of the iterative algorithm in panel (a) of \cref{fig:ista_r} as a single layer in a feedforward network. Instead of using the known matrix $\v A$ we replace the matrices in panel (a) with general matrices $\v P_1$ and $\v P_2$ to be learned, as well as the regularization parameter $\lambda$, as illustrated in panel (b) of \cref{fig:ista_r}. Thus, a single layer of this unfolded network is described by 
\begin{equation*}
\v x^{k+1}=\Tscr_{\lambda/L_f}\left\{\v P_2\v x^{k}+\v P_1\v y\right\}.
\end{equation*}
By concatenating several such layers (typically less than ten layers, corresponding to ten iterations are sufficient), a deep network is formed. 

\section{ResNet architecture}
\label{Sec:SI_re}
In this section we provide an additional result of ResNet applied to the simulated data (trained for 10 epochs on simulated data), as well as a detailed description of the complex ResNet architecture. 

\begin{figure}[!h] 
    \centering
    \includegraphics[width=1\linewidth]{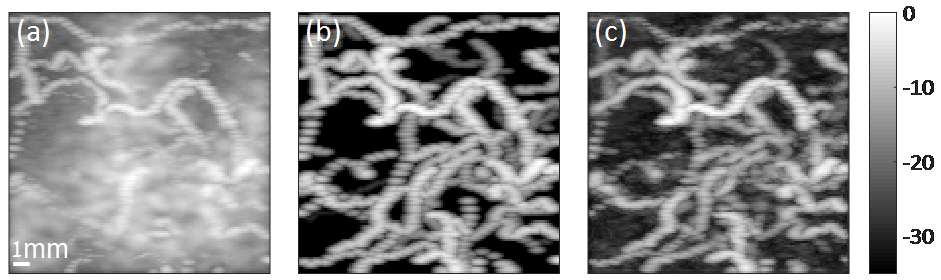}
    \caption{Simulation results of ResNet. (a) MIP image of the same simulated movie used in the main paper. (b) Ground truth MIP image of the UCAs. (c) MIP image of the UCAs recovered by ResNet. Color bar is in dB.}\label{Fig:ResNet_sim_SI}
\end{figure}

\begin{figure}[!h] 
    \centering
    \includegraphics[width=0.9\linewidth]{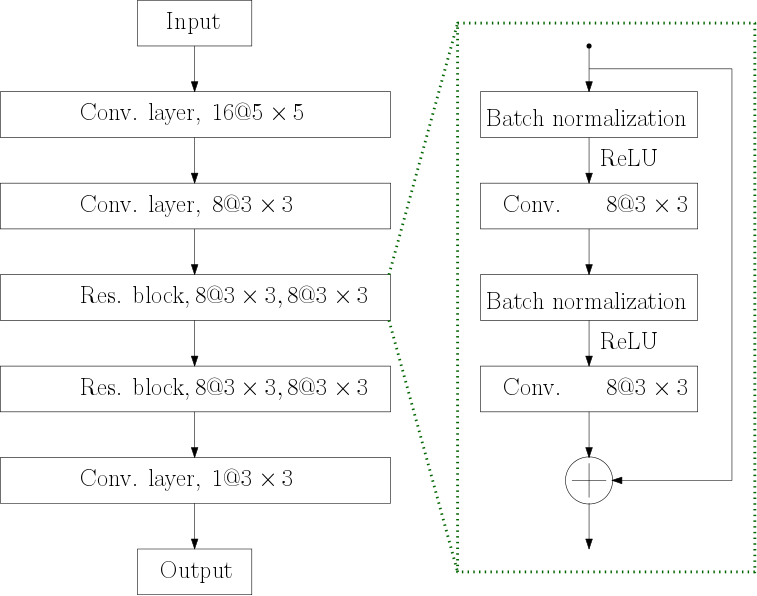}
    \caption{ResNet architecture used in this work. Conv. layers are complex convolution layers.}\label{Fig:ResNet_arch}
\end{figure}

Figure \ref{Fig:ResNet_sim_SI} presents the ResNet recovery of the same simulated movie presented in \cref{fig:sim_unfolded} of the main paper. Visual inspection of \cref{fig:sim_unfolded} reveals that in the case of simulations, ResNet suppresses the tissue signal and reveals the UCA signal. However, comparing panel (c) of \cref{Fig:ResNet_sim_SI} to panel (c) of \cref{fig:sim_unfolded} of the main paper shows that the CORONA reconstruction achieves higher contrast, in line with the conclusions drawn in the main paper. Moreover, CORONA is able to recover the tissue signal as well as the UCA signal, whereas ResNet recovers the UCA signal only. 
Figure \ref{Fig:ResNet_arch} shows the ResNet architecture used in this work. Here, Conv. layer is a complex convolution layer, and $16@5\times 5$ refers to 16 convolution channels with a $5\times 5$ pixels kernel. 

\section{Training loss functions and learned regularization parameters}
\label{Sec:SI_losses}
In this section, we provide the training and validation losses for the training process of the unfolded network and ResNet. Training was performed in two stages. The first stage consisted of 50 training epochs over 2400 simulated movie patches (20 frames each), while the second stage included additional 20 training epochs over 2400 patches from the first rat (20 frames each). For {\it in-vivo} validation, 100 consecutive frames from the second rat were chosen randomly. MSE was calculated according to (8) in the main paper.

\begin{figure*}[!h]%
\centering
\subfigure[Training on simulation data only.]{%
\label{fig:loss_sim}%
\includegraphics[height=2.65in]{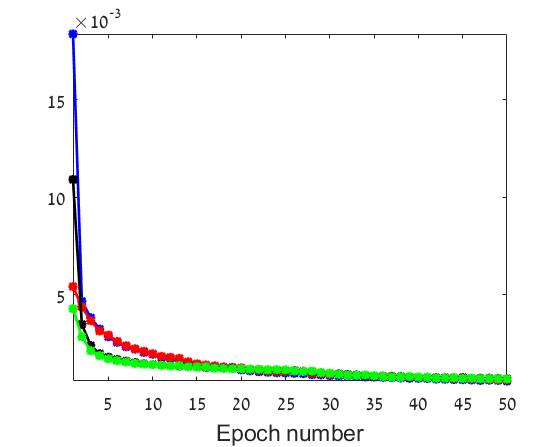}}
\qquad
\subfigure[Training on simulation and {\it in-vivo} data.]{%
\label{fig:loss_invivo}%
\includegraphics[height=2.65in]{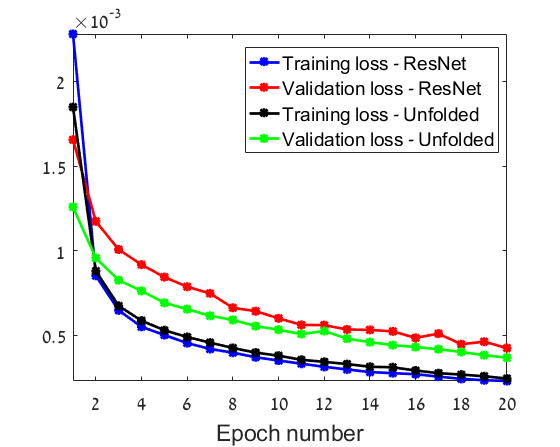}}
\caption{Training and validation losses for the unfolded network and ResNet. Left panel shows both training and validation losses for training the unfolded network (10 layers) and ResNet with simulation patches only for 50 epochs. Right panel presents both training and validation losses for the same networks trained with simulation patches for 50 epochs and additional 20 epochs on {\it in-vivo} data.}\label{Fig:losses}
\end{figure*}

Considering \cref{Fig:losses}, it is evident that when training on simulation data only, the validation curves follow the training loss curves for both networks and are comparable after 20 epochs. This behavior might suggest that the networks over-fit the simulated data, that is, they achieve the best possible recovery for simulated patches. However, in this case, the networks have yet to learn from actual data. In such a case, if the simulation does not represent the data precisely (e.g. different dynamic range, MB concentration, etc.), its performance will degrade when applied to {\it in-vivo data}, as presented in \cref{Sec:SI_TraningImportance}. Thus, additional training is performed, as shown in panel (b) of \cref{Fig:losses}. In this case, the validation losses are higher than the training loss, but now, as presented in the main paper, the networks perform well on {\it in-vivo} data.

\begin{figure}[ht] 
    \centering
    \hspace{-2.3cm}\includegraphics[width=1.1\linewidth]{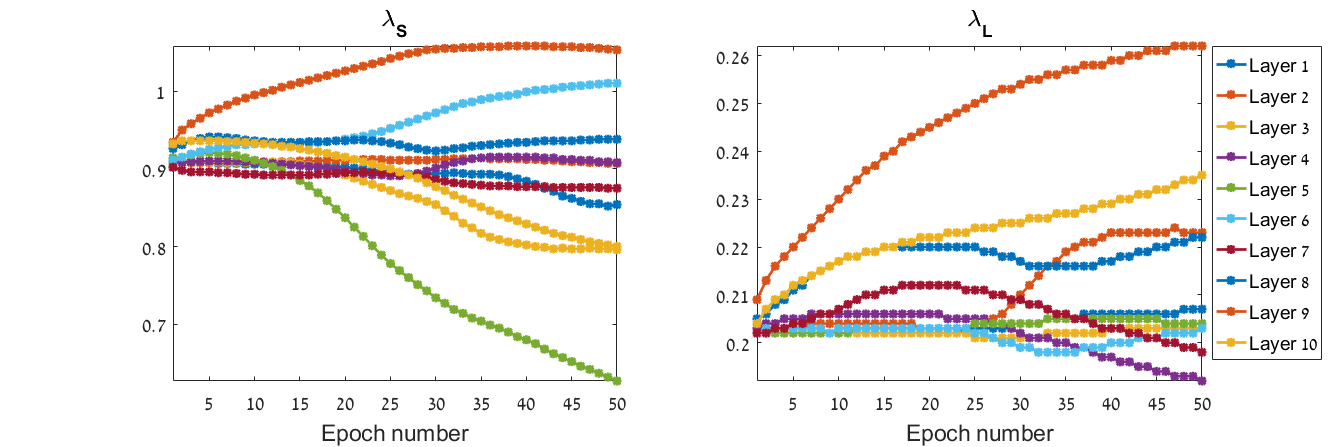}
    \caption{Learned regularization parameters when training on simulation data only.}\label{fig:lambdas_sim}
\end{figure}

\begin{figure}[ht] 
    \centering
    \hspace{-2.3cm}\includegraphics[width=1.1\linewidth]{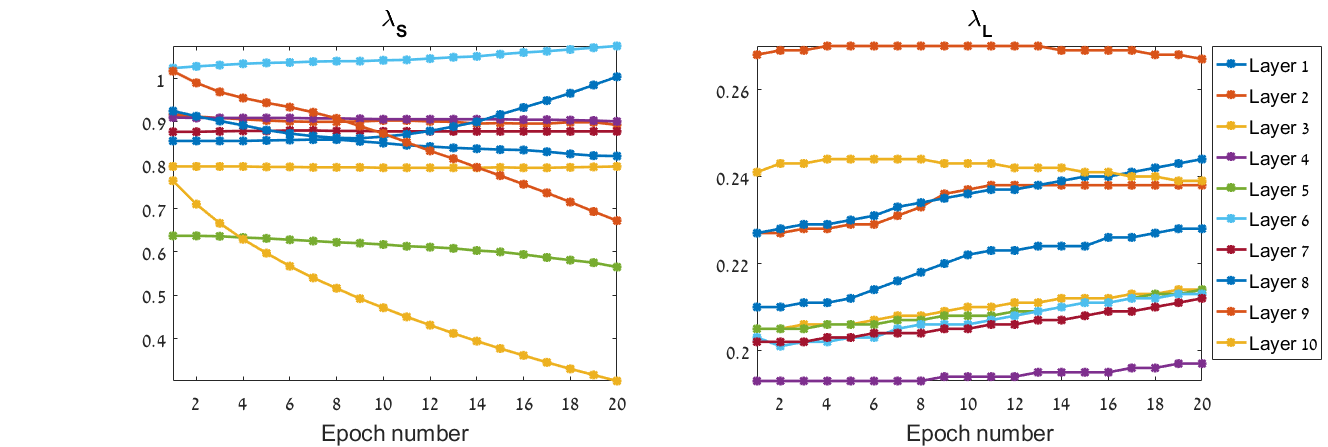}
    \caption{Learned regularization parameters when training on simulation and {\it in-vivo} data.}\label{fig:lambdas}
\end{figure}

Figure \ref{fig:lambdas_sim} and \cref{fig:lambdas} illustrate the learned values of $\lambda_{L_i}$ and $\lambda_{S_i}$ for the unfolded network, where $i=1,\ldots,10$ indicates the layer number when training on simulation data only and on simulation and {\it in-vivo} data together.

Considering both figures, it is evident that most of the regularization parameters do not change considerably when training on {\it in-vivo} data is performed. As the unfolded network is trained on both simulation and {\it in-vivo} data, the regularization parameters do not converge to the parameters used in the iterative FISTA algorithm. This also suggests that by performing combined learning on both simulation and experimental data, the network further differs from its iterative counterpart, often leading to improved performance, as presented in the main paper.

\section{The importance of training on both simulations and in-vivo data}
\label{Sec:SI_TraningImportance}
As was described in the main paper and in \cref{Sec:SI_losses}, the unfolded network outperforms FISTA reconstruction due to the combined training on both simulations and {\it in-vivo} data. This joint training allows the network to learn both the "ideal conditions" for MB/tissue separation from the simulations, as well as important features from the experimental data, and achieve robustness to noise and modeling mismatch. 

\begin{figure}[ht] 
    \centering
    \includegraphics[width=1\linewidth]{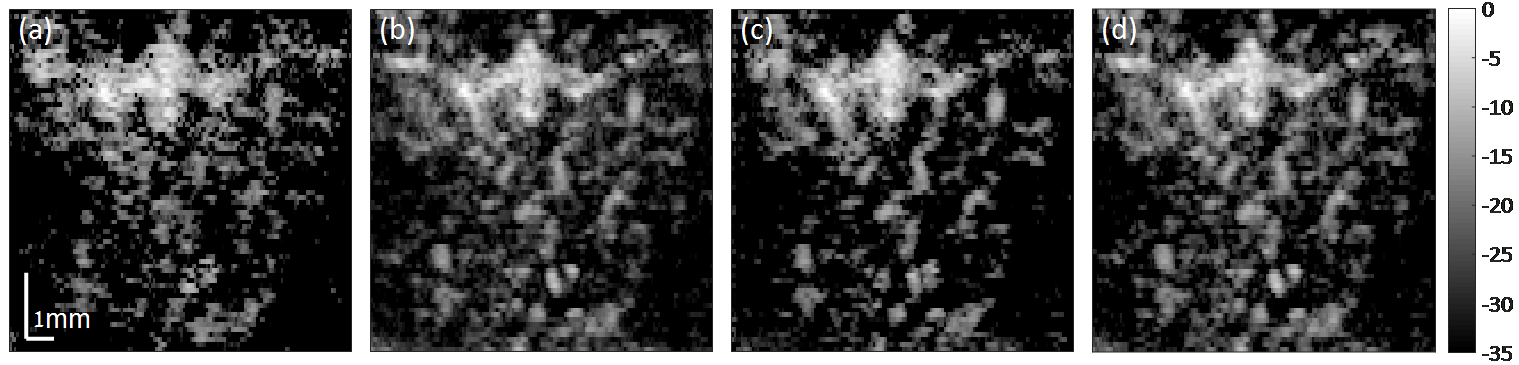}
    \caption{{\it In-vivo} results when training on simulations only and on simulations and {\it in-vivo} data together. (a) Training on simulations for 10 epochs. (b) Training on both simulations and experimental data for 10 epochs each. (c) Training on simulations for 50 epochs. (d) Training on both simulations and experimental data for 50 and 20 epochs, respectively. Color bar is in dB.}\label{fig:compare_train}
\end{figure}

In \cref{fig:compare_train} we present {\it in-vivo} results of the network trained in two conditions. Panels (a) and (b) show the output of the network when trained solely on simulated data for 10 epochs and the output when trained on both simulated and experimental data for 10 epochs each, respectively. Panels (c) and (d) show the output of the networks for the same two cases, only now the numbers of training epochs were 50 for simulated data and 20 for {\it in-vivo} data.

Considering \cref{fig:compare_train}, clearly when training on a relatively low number of epochs (10), simulated data is not sufficient for good performance on experimental data. On the other hand, when combined with additional 10 epochs of training on {\it in-vivo} data, the performance of the network improves considerably, and is somewhat similar to the performance of the network result displayed in panel (d). Surprisingly, even when training on simulated data only for enough epochs, in this case 50, the network performs well in recovering the vascular bed of experimental data, as shown in panel (c). However, closer examination shows that albeit the image looks sparser than the image in panel (d), its texture looks more pixel-like than the FISTA and SVD images shown in \cref{Fig:main} of the main paper.  

The latter example suggests two things. First, that good results can be obtained by training the network on realistic simulations for enough training epochs. The second is that performance more similar in texture and visual quality to that of non-learning based techniques can be obtained by the combined training on both simulations and experimental data.

\section{Runtime comparison}
\label{Sec:SI_runtime}
\begin{figure*}[!h]%
\centering
\subfigure[Unfolded network, 1 layer]{%
\label{fig:reuntime_unf1}%
\includegraphics[height=1.7in]{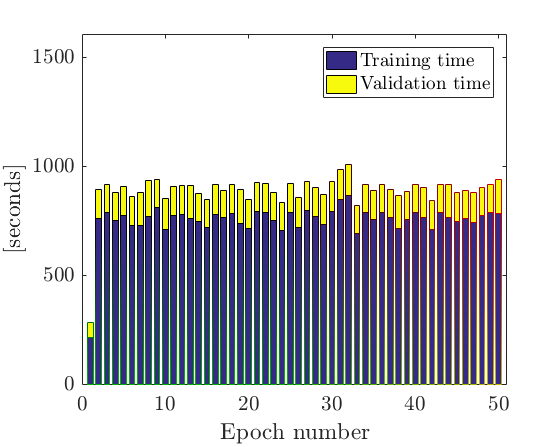}}%
\qquad
\subfigure[Unfolded network, 10 layers]{%
\label{fig:reuntime_unf10}%
\includegraphics[height=1.7in]{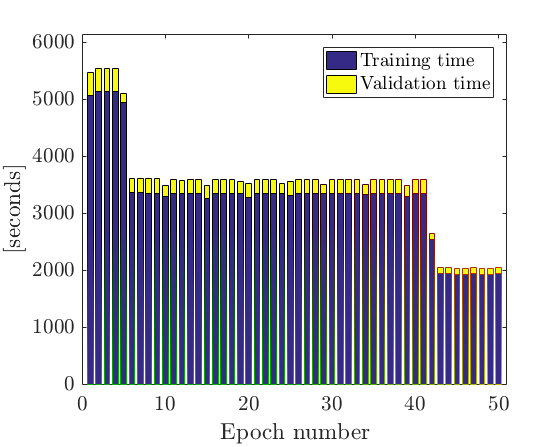}}%
\qquad
\subfigure[ResNet]{%
\label{fig:reuntime_resnet}%
\includegraphics[height=1.7in]{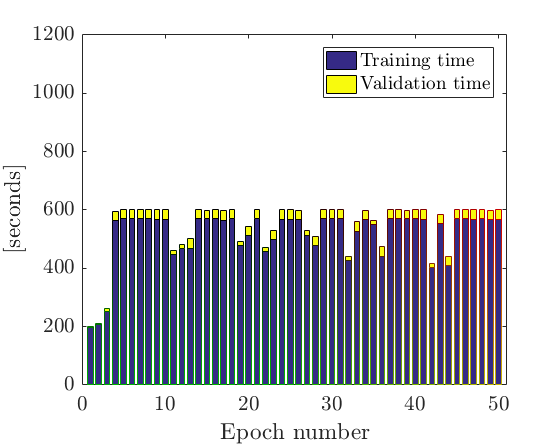}}%
\caption{Run-time results for training and validation of the unfolded network and ResNet.}\label{Fig:runtime}
\end{figure*}

Here, we compare the run-time performance of both the unfolded network and ResNet, for both training phase and validation phase. Fig. \ref{Fig:runtime} show the time in seconds each network required to train a single epoch and then validate its performance, in yellow, as a function of epoch number. Training was performed for 50 epochs on simulated data. Run-times results for training on {\it in-vivo} data were similar, and thus are omitted. 

Observing \cref{Fig:runtime}, it is evident that the training and validation of the unfolded network is slower compared with ResNet. The 10 layers network is slower by an order of magnitude, but the 1 layer network has a slightly slower runtime. 
The slower processing and training time of the unfolded network is attributed to the SVD operations required by the network, although faster and more efficient algorithms for SVD computations can be used, as discussed in the Discussion section of the main paper. This figure further supports the conclusions in the main paper. The unfolded network offers a flexible trade-off between execution time and performance, by allowing to choose its depth.

However, the unfolded network has an order of magnitude lower number of trainable parameters and achieves better CNR and CR values, as demonstrated in the main paper. It is also important to remember that the ResNet was not fully trained, rather only its last fully connected layers were trained. This transfer learning process considerably reduces the overall training time.

\section{Simulation description}\label{Sec:Sims_SI}
As was indicated in the main paper, in this work we increase the number of training examples by training on both experimental and simulated data. In this section, we describe how the simulation was generated. In the simulations we used, pixel size is assumed to be $0.12\times 0.12\textrm{mm}^2$ and the number of pixels is $128\times 128$. Implementation was performed in Python 3.5.2.

\subsection{MB signal generation}
The overall number of MBs (as well as their initial positions) was generated randomly up to a maximum concentration of $130$ MBs per $\textrm{cm}^{-2}$. MB amplitudes were drawn from a normal complex distribution. 

MB velocity magnitudes were generated according to 
\begin{equation*}
v(x,y,t=0)=\max(0, v_{\textrm{det}}\cdot\mathcal{N}(1,1)),
\end{equation*}
where $v_{\textrm{det}}=0.24\textrm{mm}/dt$, $dt=0.01$s is the imaging frame-rate and $\mathcal{N}(1,1)$ is a normal distribution with mean 1 and standard deviation 1. MB accelerations were generated according to 
\begin{equation*}
a_{x/y}=\mathcal{N}(0,\sigma_a),
\end{equation*}
with $\sigma_a=0.05\cdot 0.12/dt^2$ and $x/y$ are the lateral and axial directions, respectively. 

MB velocity directions are generated in each frame according to 
\begin{equation*}
v_x^{k}(t)=v_x^{k-1}(t)\textrm{cos}(\theta)-v_y^{k-1}(t)\textrm{sin}(\theta),
\end{equation*}
\begin{equation*}
v_y^{k}(t)=v_x^{k-1}(t)\textrm{sin}(\theta)+v_y^{k-1}(t)\textrm{cos}(\theta),
\end{equation*}
with $\theta\sim \textrm{U}[-30^\circ,30^\circ]$ and $k$ indicates frame number. MB amplitudes are additionally multiplied by a random factor between 0.9 and 1.1 in each frame.

\subsection{Tissue signal generation}
To model the tissue signal, we start by generating a sum of five real 2D Gaussian matrices of the same size as the image frames ($128\times 128$ pixels) with random positions and variances. We then generate a complex random matrix to modulate the envelope of the tissue signal. The real and complex entries are both drawn from a normal distribution with zero mean and standard deviation 1. Both matrices are then multiplied element-wise, and the product is then low-pass filtered (2D real Gaussian matrix of $11\times 11$ pixels). The resulting signal's envelope, denoted as $\v B\in\mathcal{R}^{I\times J}$ mimics the texture of the tissue signal. Thus, the overall pixels' values are random, but locally they are correlated. 

The next step involves the generation of a phase matrix, same size as before.  It's entries are drawn from a Gaussian distribution in the following manner 
\begin{equation*}
{\bm \theta}\sim \mathcal{N}(\alpha, \sigma_{\theta}),
\end{equation*}
with a mean drawn from a uniform distribution in the range $\alpha\sim[0^\circ,180^\circ]$ and standard deviation of $	\sigma_{\theta}=15^\circ$. The resulting complex tissue signal is given by
\begin{equation*}
\v T[i,j]=\v B[i,j]e^{j{\bm\theta}[i,j]},\;i,j\in[I,J].
\end{equation*}

The next stage in generating the tissue signal is to apply spatial deformations in each frame, to mimic tissue movement during the acquisition period. To this end we start by generating four different $4\times 4$ kernels, denoted as flow filters. The entries of those kernels, are positive and their sum equals to one. For each new frame, we generate additional four filters, with entries drawn from a Gaussian distribution with zero mean and standard deviation of 0.1. For each flow filter we add the corresponding new filter and for each pixel we take the maximum between the latter value and 0.1. The resulting new filter is normalized such that all entries sum to one and the entries are non-negative. 

Once the flow filters have been updated for the current frame, they are convolved with $\v T$, to get four different images. We then divide each of the four images into $4\times 4$ blocks. The final image is generated by dividing an empty matrix into $4\times 4$ blocks, and for each block choosing randomly one of the corresponding blocks from the four images. This process ensures that blocks in the same neighborhood share the same movement pattern, but in the whole image, the pattern is random.

\subsection{Simulation of the PSF}
Once the (complex) MB frame and  tissue frame are generated and summed with complex Gaussian noise, the resulting frame is convolved with the PSF. The PSF is modeled as a 2D real 
Gaussian kernel with standard deviations of $0.14$mm in the lateral and $0.32$mm axial dimensions, taken from the {\it in-vivo} data. 
\end{document}